\documentclass[letterpaper, 10 pt, conference]{ieeeconf}  %

\usepackage{cite}
\usepackage{amsmath,amssymb,amsfonts}
\usepackage{graphicx}
\usepackage{textcomp}
\usepackage[table]{xcolor}
\usepackage{booktabs}
\usepackage{multirow}
\usepackage{subcaption}
\usepackage{hyperref}
\usepackage{url}
\usepackage{dsfont}

\usepackage{fancyhdr}
\fancypagestyle{withfooter}{
  \fancyhf{} %

  \fancyfoot[C]{\footnotesize Accepted to the IEEE ICRA Workshop on Open Challenges for Rigorous Robot Perception 2026}
}

\IEEEoverridecommandlockouts                              %

\title{\LARGE \bf
Spatially Stratified Distillation for  \\ Heterogeneous Radar  Place Recognition}

\author{Sagun Singh Shrestha$^{1}$, Samuel Harding$^{2}$, Abdelwahed Khamis$^{1}$, Saimunur Rahman$^{1}$, Peyman Moghadam$^{1,2}$ %
\thanks{$^{1}$ CSIRO Robotics, Brisbane, Australia}%
\thanks{$^{2}$ University of Queensland, Brisbane, Australia}%
}

\begin{document}

\maketitle

\thispagestyle{withfooter}
\pagestyle{withfooter}

\begin{abstract}

Scalable, all-weather place recognition increasingly relies on heterogeneous radar place recognition to bridge diverse hardware platforms. A notable application is matching queries from cost-effective 4D automotive radars against high-fidelity reference maps built by dense spinning radars. This process is fundamentally limited by the extreme sparsity (and narrow field-of-view)  of the 4D sensor, which captures only a fraction of the structural density present in the spinning radar database. Prior efforts address this issue by unifying different radar signals. That is, projecting both signals into a common representational space. Yet, they suffer performance degradation in multi-session environments. In this paper, we propose spatially-stratified distillation (SSD); a strategy that replaces standard uniform distillation with an asymmetric spatial alignment derived directly from physical radar returns. In regions where both radars exhibit overlapping returns, SSD enforces strong feature alignment. Crucially, in sparse regions where the 4D student lacks returns but the teacher contains valid structure within the shared field of view, SSD applies heavily discounted distillation weights. 
Extensive evaluations of the recent HeRCULES dataset demonstrate that SSD significantly outperforms prior place recognition methods, achieving state-of-the-art results on its challenging dynamic sequences.
\end{abstract}

\vspace{0.5em}
\vspace{0.5em}

\section{Introduction}

\begin{figure}[t]
    \centering
    \includegraphics[width=1\linewidth]{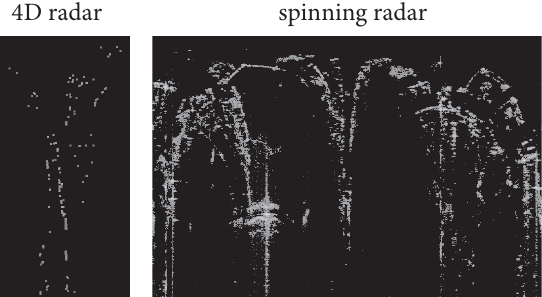}
    \caption{Radar place recognition challenge addressed in this paper via SSD. The 4D radar view on the left is a sparse, narrow-field observation of the same scene shown by the spinning radar on the right. \textit{Question: Can you identify where the 4D radar view fits inside the spinning-radar scan?} The answer is revealed on the next page.}
    \label{fig:radar-gap}
    \vspace{-2em}
\end{figure}

Robust autonomous navigation relies on place recognition \cite{vidanapathirana2022logg3d,vidanapathirana2023spectral,hausler2025pair,knights2026wildcross}. Radar sensors are highly attractive for this task due to their immunity to adverse weather and lighting conditions that often compromise cameras and LiDAR. While high-fidelity environmental reference maps are typically constructed using expensive, dense $360^\circ$ spinning radars, mass-market vehicles are increasingly equipped with cost-effective, solid-state 4D automotive radars. Consequently, scalable localization requires \textit{heterogeneous radar place recognition}: matching sparse 4D radar queries against dense spinning radar databases. However, this process is fundamentally bottlenecked by severe modality asymmetry. As shown in Fig.~\ref{fig:radar-gap}, the 4D radar sensor captures a sparse $120^\circ$ view of the scene, whereas the spinning radar captures a dense $360^\circ$ panorama of the entire scene.

SHeRLoc~\cite{kim2025sherloc}, a recent work, bridges this cross-modal gap by projecting both data types into a synchronized polar representation and extracting features via a shared backbone. While effective for single-session retrieval, multi-session retrieval,  where the map and query are recorded on different days along the same route, remains challenging (Sec.~\ref{sec:muli_session_eval}). The principal source of difficulty is the severe sparsity of the 4D radar queries compared to the dense spinning radar database: a shared-backbone metric learning based solution naturally aligns features where both sensors observe identical structures, but it does not attempt to leverage the dense structural context present in the spinning radar map that is physically absent from the 4D radar query.

To bridge this gap, we propose \textit{Spatially-Stratified Distillation} (SSD), which reformulates cross-modal distillation specifically for asymmetric sensing. SSD operates through three core mechanisms designed to inject dense structural supervision into the sparse 4D student without violating physical sensing constraints.
Specifically, we derive a distillation objective that strongly aligns jointly observed regions while utilizing unobserved, teacher-only regions as a heavily discounted structural prior. This elevates the 4D radar student learned representation while at the same time preventing it from hallucinating unobservable structures.

By targeting the regions where sparsity is worst, SSD provides an external teaching signal exactly where the student needs it most, significantly improving single-session and multi-session retrieval performance without adding complexity at inference time. Our core contributions are:
\begin{enumerate}

    \item \textbf{Spatially-Stratified Distillation (SSD):} We propose a cross-modal framework bridging sparse 4D and dense spinning radars. SSD employs an \emph{asymmetric physical mask} to leverage dense teacher context without forcing the student to hallucinate unobservable structures.

    \item \textbf{State-of-the-art Performance:} On the HeRCULES benchmark, SSD achives state-of-the-art and  outperforms best performing baselines \cite{kim2025sherloc} on all single-session and multi-session sequences in heterogeneous (4D query \emph{vs.} spinning database, AR@1 $0.766$ \emph{vs.} $0.722$) and homogeneous (4D query \emph{vs.} 4D database, AR@1 $0.954$ vs $0.905$) settings.
\end{enumerate}

\begin{figure}[t!]
    \centering
    \includegraphics[width=1\linewidth]{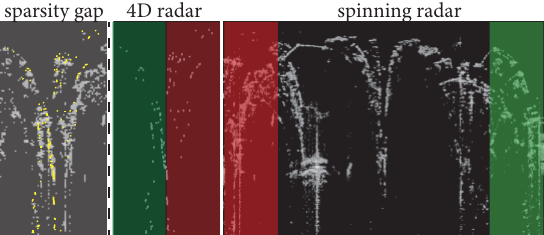}
    \caption{\textbf{Right:} Highlighted region locating the sparse 4D radar view within the dense spinning-radar scan. \textbf{Left:} 4D radar returns (yellow colored) overlaid on the spinning-radar region. Although acquired simultaneously from the same scene, the modalities exhibit a disparity in measurement density in the polar RCS representation, where horizontal and vertical axes denote azimuth and range, respectively.}
    \label{fig:solution}
\end{figure}

\section{Proposed Method}
\label{sec:method}

\textbf{\textit{Notation.}} We use $s$ for the student branch operating on 4D radar and $t$ for the teacher branch operating on spinning radar. Let
$\mathbf{I}_s \in \mathbb{R}^{1 \times H \times W}$ and
$\mathbf{I}_t \in \mathbb{R}^{1 \times H \times W}$ be the polar BEV inputs,
and let
$\phi_s, \phi_t : \mathbb{R}^{1 \times H \times W}
                 \!\to\! \mathbb{R}^{C \times H_f \times W_f}$
be the corresponding backbones, producing feature maps
$\mathbf{F}_s = \phi_s(\mathbf{I}_s)$ and
$\mathbf{F}_t = \phi_t(\mathbf{I}_t)$
at stride $r = H/H_f$. We index spatial positions by
$\mathbf{p} = (u,v) \in \Omega := \{1,\dots,H_f\} \times \{1,\dots,W_f\}$
and write $\mathbf{F}[\mathbf{p}] \in \mathbb{R}^C$ for the channel vector at
$\mathbf{p}$. The per-location unit vector is
$\hat{\mathbf{F}}[\mathbf{p}] := \mathbf{F}[\mathbf{p}] \,/\,
\|\mathbf{F}[\mathbf{p}]\|_2$.
We use $\mathds{1}[\cdot]$ for the indicator function, $G_\sigma$ for a
2D Gaussian kernel of bandwidth $\sigma$, and $\ast$ for 2D convolution.
Tensors are written for a single sample throughout.

\textbf{\textit{Base pipeline.}}
Given a 4D-radar query frame and a spinning-radar database frame, polar BEV inputs $\mathbf{I}_s, \mathbf{I}_t \in \mathbb{R}^{1\times H\times W}$  are produced with a matching $120^\circ$ FOV following ~\cite{kim2025sherloc}. The feature extraction backbones produce local feature maps $\mathbf{F}_s = \phi_s(\mathbf{I}_s)$, $\mathbf{F}_t = \phi_t(\mathbf{I}_t) \in \mathbb{R}^{C\times H_f\times W_f}$
with $C=256$ and $(H_f, W_f)\approx(16,80)$ after 8$\times$ downsampling. Optimal transport based HOLMES pooling ~\cite{kim2025sherloc} was used to aggregate each local map to a global descriptor $\mathbf{d}\in\mathbb{R}^D$ with $D=320$. Training uses FOV-aware adaptive-margin triplet loss $\mathcal{L}_{\mathrm{trip}}(\mathbf{d}_q, \mathbf{d}_p, \mathbf{d}_n)$ \cite{kim2025sherloc}, where the query $\mathbf{d}_q$ comes from the student branch (4D radar) and the positives \& negatives $\mathbf{d}_p, \mathbf{d}_n$ come from the teacher branch (spinning radar).

 \begin{figure*}
     \centering
     \includegraphics[width=0.8\linewidth]{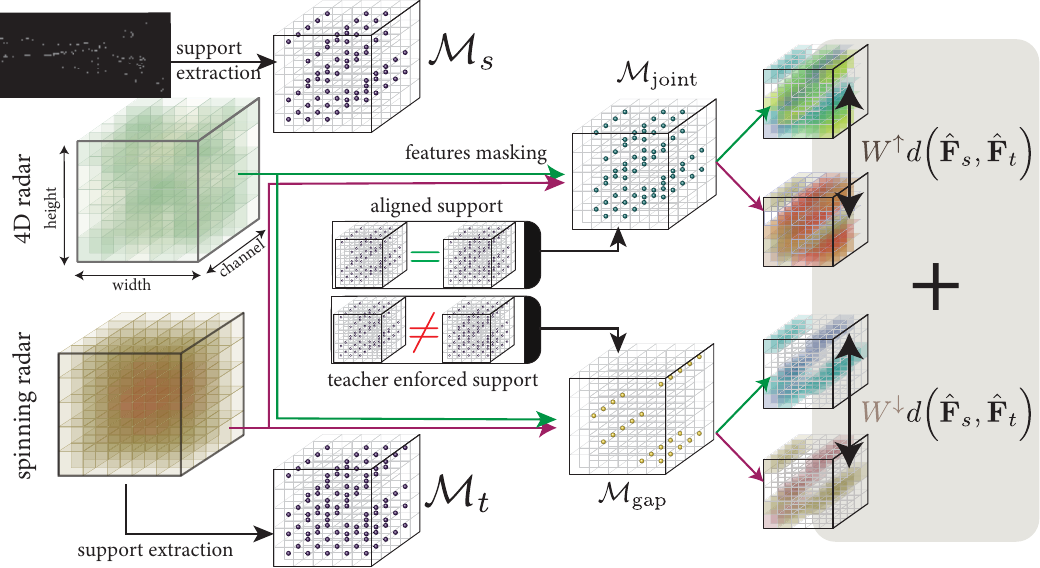}
     \caption{\textbf{Overview of Spatially-Stratified Distillation (SSD).} The distillation pipeline explicitly accounts for the field-of-view asymmetry between the sparse 4D radar student and the dense spinning radar teacher. First, support extraction isolates physical measurements ($\mathcal{M}_s$) and meaningful teacher activity ($\mathcal{M}_t$). These masks are combined to stratify the spatial domain into two regimes. In regions of aligned support ($\mathcal{M}_{\mathrm{joint}}$), the feature alignment loss is strictly enforced with a high weight ($W^\uparrow$). Conversely, teacher-supported locations falling into the student's unobserved field-of-view ($\mathcal{M}_{\mathrm{gap}}$) are not discarded; they are distilled with a heavily discounted weight ($W^\downarrow$) to serve as a weak structural prior, ensuring the student is guided by global context without being penalized for physical sensing limitations.}
     \label{fig:str_distill}
    \vspace{-1.5em}
 \end{figure*}

As illustrated in Fig.~\ref{fig:str_distill}, we use the dense spinning radar teacher to enrich the learned 4D features. However, the teacher captures contextual cues physically unavailable to the sparse 4D student. This informational disparity makes exact feature alignment infeasible. Instead, we treat the detached teacher as a structural prior~\cite{bang2024radardistill}, aligning intermediate activations via:
\begin{equation}
\mathcal{L}_{\mathrm{FD}}
=
\frac{
\sum_{\mathbf{p}\in\Omega}
W(\mathbf{p})\,
d\!\left(\hat{\mathbf{F}}_s[\mathbf{p}],\hat{\mathbf{F}}_t[\mathbf{p}]\right)
}{
\sum_{\mathbf{p}\in\Omega} W(\mathbf{p}) + \varepsilon
},
\label{eq:fd_general}
\end{equation}
where the spatial mask $W(\mathbf{p})$ dictates the distillation weight and $d(\cdot,\cdot)$ measures feature discrepancy. Crucially, applying a uniform spatial mask is physically invalid due to the severe field-of-view mismatch. Uniform alignment forces the student to hallucinate unobservable structures, whereas completely masking unobserved regions discards the dense contextual prior.
On the other hand, masking out unobserved regions entirely would discard the contextual prior that makes distillation valuable in the first place. 

To resolve this asymmetry, we propose \textbf{Spatially-Stratified Distillation (SSD)}, which reformulates the distillation objective to account for both the sensing geometry and the modality differences of the problem. Specifically, SSD introduces a spatial weighting function $W(\mathbf{p})$ that distinguishes between regions of joint observability and regions unique to the teacher:
\begin{equation}
W(\mathbf{p}) = w_h\,\mathcal{M}_{\mathrm{joint}}(\mathbf{p}) + w_\ell\,\mathcal{M}_{\mathrm{gap}}(\mathbf{p}),
\label{eq:afd_weight}
\end{equation}
where $ w_\ell\ll w_h$. $\mathcal{M}_{\mathrm{joint}}$ represents the \textit{Joint Observations Mask} where both the student and teacher have physical measurement support. $\mathcal{M}_{\mathrm{gap}}$ is the \textit{Unobserved Context Mask}. This mask isolates locations where the teacher contains meaningful structural information ($\mathcal{M}_t$) but the student lacks sufficient physical support (student evidence is below a threshold $\tau_s$). This weighting is central to our formulation. 
It preserves strong supervision where spatial correspondence is physically credible, while using teacher-only regions as a weak regularizing prior rather than as a hard matching target. The masks are leveraged as follows:
 
\begin{equation}
\mathcal{M}_{\mathrm{joint}}(\mathbf{p}) = \mathcal{M}_s(\mathbf{p})\,\mathcal{M}_t(\mathbf{p}).
\label{eq:mhigh}
\end{equation}
\vspace{-1em}
\begin{equation}
\mathcal{M}_{\mathrm{gap}}(\mathbf{p}) = \bigl(1-\mathds{1}[\mathcal{M}_s(\mathbf{p})>\tau_s]\bigr)\,\mathcal{M}_t(\mathbf{p}), 
\label{eq:mlow}
\end{equation}

Importantly, $\mathcal{M}_{\mathrm{gap}}$ does not defy physical plausibility; it simply captures teacher-supported locations that \textbf{fall into the student's FoV gap} without forcing an impossible exact match. Rather than discarding these locations, we incorporate them with a discounted weight. The mechanism also ensures that the student is never penalized as if it had direct access to unobserved content, while remaining weakly regularized toward representations compatible with the teacher's richer global scene structure.

\textbf{\textit{Mask Generation.}}
To effectively isolate these regions, the underlying masks ($\mathcal{M}_t$ and $\mathcal{M}_s$) must accurately reflect the distinct informational roles of each branch. 
For the teacher branch, which processes a dense scene representation, we use a data-adaptive magnitude mask applied directly to the teacher's feature maps: $\mathcal{M}_t(\mathbf{p}) = \mathds{1}\!\left[\|\mathbf{F}_t[\mathbf{p}]\|_2 > \alpha \,\bar{F}_t \right]$
where $\bar{F}_t = \frac{1}{|\Omega|} \sum_{\mathbf{p}'\in\Omega} \|\mathbf{F}_t[\mathbf{p}']\|_2$ is the spatial feature magnitude.

Unlike prior works~\cite{bang2024radardistill} that rely on intermediate features, which decouple from physical evidence due to convolutional diffusion, we derive the student mask directly from the downsampled 4D input $\mathbf{I}_s^\downarrow$ as $\mathcal{M}_s(\mathbf{p}) = \left(G_\sigma * \mathds{1}[\mathbf{I}_s^\downarrow > 0]\right)(\mathbf{p}),$
where $G_\sigma$ is a Gaussian kernel.

\textbf{\textit{Activation alignment.}} Finally, to address activation scale and correlation mismatches between branches, we align the student channel basis using a $1\times1$ convolution $\psi$ and apply $L_2$-normalization. We supervise the normalized features using cosine distance, $d(\mathbf{u},\mathbf{v}) = 1 - \langle \mathbf{u},\mathbf{v}\rangle$, and further align their second-order structure through
$\mathcal{L}_{\mathrm{cov}} = \|\operatorname{cov}(\mathbf{u})-\operatorname{cov}(\mathbf{v})\|_F^2$.
The resulting feature-distillation loss incorporates the spatial weighting $W(\mathbf{p})$ into the general objective~\eqref{eq:fd_general} and is optimized jointly with the primary place-recognition triplet loss:
\begin{equation}
\mathcal{L} = \mathcal{L}_{\mathrm{trip}}
+ \lambda_{\mathrm{FD}}\mathcal{L}_{\mathrm{FD}}
+ \lambda_{\mathrm{cov}}\mathcal{L}_{\mathrm{cov}},
\label{eq:total_loss}
\end{equation}
where $\lambda_{\mathrm{FD}}$ and $\lambda_{\mathrm{cov}}$ denote the weights of the feature-distillation and covariance-alignment losses, respectively.

\section{Experimental Evaluation}
\label{sec:exp}

We evaluate SSD on the HeRCULES dataset \cite{kim2025hercules}, following the SHeRLoc protocol \cite{kim2025sherloc}. Training is performed using \textit{Mountain 01–03}, \textit{Bridge 01}, \textit{Stream 02}, and \textit{Parking Lot 03–04} sequences, while evaluation is done on \textit{Sports Complex 01–03}, \textit{Library 01–03}, and \textit{River Island 01-02} sequences in single-session and multi-session settings. Performance is measured using Recall@K (R@K) and Average Recall@K (AR@K), with predictions considered correct if they fall within 5 m of the ground truth. Our method is implemented in PyTorch, using a ResNet based backbone similar to SHeRLoc, and also trained for 20 epochs on multiple GPUs.

\subsection{Heterogeneous Single-session Place Recognition}
Tab. \ref{tab:sota} shows the results of SSD on single-session place recognition. In this evaluation, we use spinning radar data as the database and 4D radar data as queries across the Sports
Complex, Library, and River Island sequences. The results show that SSD consistently surpasses all other methods including the very recent state-of-art baselines.

\begin{table*}[t]
\caption{Single-session place recognition on HeRCULES dataset. 
SSD results are mean over three runs.}
\label{tab:sota}
\centering
\scriptsize
\setlength{\tabcolsep}{3pt}
\renewcommand{\arraystretch}{1.1}
\begin{tabular}{l | cc cc cc | cc cc cc | cc cc cc}
\toprule
\multirow{4}{*}{\textbf{Methods}}
& \multicolumn{6}{c|}{\textbf{Sports Complex}}
& \multicolumn{6}{c|}{\textbf{Library}}
& \multicolumn{6}{c}{\textbf{River Island}} \\
& \multicolumn{2}{c}{01}
& \multicolumn{2}{c}{02}
& \multicolumn{2}{c|}{03}
& \multicolumn{2}{c}{01}
& \multicolumn{2}{c}{02}
& \multicolumn{2}{c|}{03}
& \multicolumn{2}{c}{01}
& \multicolumn{2}{c}{02}
& \multicolumn{2}{c}{03} \\
\cmidrule{2-7}\cmidrule{8-13}\cmidrule{14-19}
& R@1 & R@1\%
& R@1 & R@1\%
& R@1 & R@1\%
& R@1 & R@1\%
& R@1 & R@1\%
& R@1 & R@1\%
& R@1 & R@1\%
& R@1 & R@1\%
& R@1 & R@1\% \\
\midrule

Radar SC~\cite{kim2020mulran}
& 0.047 & 0.135 & 0.042 & 0.061 & 0.044 & 0.064
& 0.040 & 0.060 & 0.041 & 0.063 & 0.042 & 0.066
& 0.010 & 0.020 & 0.004 & 0.012 & 0.005 & 0.030 \\

RaPlace~\cite{jang2023raplace}
& 0.042 & 0.148 & 0.056 & 0.156 & 0.025 & 0.152
& 0.035 & 0.069 & 0.040 & 0.090 & 0.016 & 0.123
& 0.002 & 0.152 & 0.000 & 0.024 & 0.004 & 0.162 \\

RadVLAD~\cite{gadd2024open}
& 0.019 & 0.334 & 0.051 & 0.222 & 0.022 & 0.426
& 0.011 & 0.237 & 0.014 & 0.295 & 0.012 & 0.245
& 0.001 & 0.217 & 0.066 & 0.543 & 0.006 & 0.367 \\

FFT-RadVLAD~\cite{gadd2024open}
& 0.040 & 0.344 & 0.014 & 0.186 & 0.022 & 0.377
& 0.023 & 0.287 & 0.014 & 0.273 & 0.011 & 0.229
& 0.002 & 0.330 & 0.026 & 0.533 & 0.001 & 0.488 \\

Autoplace~\cite{cai2022autoplace}
& 0.015 & 0.198 & 0.046 & 0.098 & 0.017 & 0.186
& 0.010 & 0.129 & 0.031 & 0.131 & 0.018 & 0.122
& 0.003 & 0.051 & 0.001 & 0.009 & 0.001 & 0.021 \\

\midrule
SHerLoc-S \cite{kim2025sherloc}
& 0.857 & 0.924 & {0.975} & 0.985 & 0.945 & {0.981}
& 0.866 & 0.917 & 0.925 & {0.950} & 0.850 & 0.891
& {0.899} & {0.965} & {0.868} & {0.963} & 0.850 & 0.945 \\

SHerLoc \cite{kim2025sherloc}
& {0.900} & {0.936} & 0.962 & {0.987} & {0.958} & 0.980
& {0.881} & {0.936} & {0.938} & {0.964} & {0.868} & {0.912}
& 0.880 & 0.957 & 0.860 & 0.952 & {0.858} & {0.959} \\

\midrule

\rowcolor{green!20}
\textbf{SSD (Ours)}
& \textbf{0.922} & \textbf{0.971} & \textbf{0.975} & \textbf{0.992} & \textbf{0.974} & \textbf{0.998}
& \textbf{0.900} & \textbf{0.968} & \textbf{0.953} & \textbf{0.980} & \textbf{0.894} & \textbf{0.966}
& \textbf{0.939} & \textbf{0.991} & \textbf{0.926} & \textbf{0.988} & \textbf{0.890} & \textbf{0.975} \\

\bottomrule
\end{tabular}
\end{table*}

\subsection{Heterogeneous Multi-session place recognition}
\label{sec:muli_session_eval}

We also evaluate SSD on multi-session evaluation. In this setting, spinning radar data from sequence 01 were
used as the database, and 4D radar data from sequences 02
and 03 were used as queries. We show the results on Tab. \ref{tab:multi}. The results show that SSD surpasses all methods including the both SHeRLoc variants in a consistent manner by a significant margin (more than 4\% over the best art).

\subsection{Homogeneous Multi-session place recognition}
Since SSD targets at improving 4D radar representation, we conduct homogeneous multi-session place recognition experiments to understand whether the learned improved representation applies beyond heterogeneous settings. Tab. \ref{tab:multi} shows the results. It is interesting to see that both Sports Complex and Library results have been improved significantly. This demonstrate that SSD indeed improved 4D representation in a favourable manner.

\begin{table}[t]
\caption{Multi-session place recognition on HeRCULES. Heterogeneous: 4D query \emph{vs} spinning database; Homogeneous-4D: 4D query \emph{vs} 4D database (student-only evaluation of the SSD-trained backbone). R@1 at 5\,m radius; R@1\% reports recall of top 1\%. AR@1 = mean over the four pairs. 
SSD results are mean over three runs.}
\label{tab:multi}
\centering
\setlength{\tabcolsep}{2pt}
\renewcommand{\arraystretch}{1.05}
\resizebox{\linewidth}{!}{%
\begin{tabular}{c l | cc cc | cc cc | c}
\toprule
 & \multirow{3}{*}{\textbf{Methods}}
 & \multicolumn{4}{c|}{\textbf{Sports Complex}}
 & \multicolumn{4}{c|}{\textbf{Library}}
 & \multirow{3}{*}{\textbf{AR@1}} \\
 &  & \multicolumn{2}{c}{01 $\rightarrow$ 02}
 & \multicolumn{2}{c|}{01 $\rightarrow$ 03}
 & \multicolumn{2}{c}{01 $\rightarrow$ 02}
 & \multicolumn{2}{c|}{01 $\rightarrow$ 03}
 &  \\
 &  &
 R@1 & R@1\% & R@1 & R@1\% &
 R@1 & R@1\% & R@1 & R@1\% &  \\
\midrule

\multirow{7}{*}{\rotatebox{90}{Heterogeneous}}
 & RaPlace~\cite{jang2023raplace}      & 0.024 & 0.056 & 0.016 & 0.057 & 0.021 & 0.044 & 0.011 & 0.023 & 0.018 \\
 & RadVLAD~\cite{gadd2024open}      & 0.011 & 0.118 & 0.005 & 0.214 & 0.012 & 0.138 & 0.011 & 0.073 & 0.010 \\
 & FFT-RadVLAD~\cite{gadd2024open}  & 0.006 & 0.121 & 0.007 & 0.182 & 0.015 & 0.217 & 0.016 & 0.136 & 0.011 \\
 & Autoplace~\cite{cai2022autoplace}  & 0.007 & 0.071 & 0.012 & 0.070 & 0.022 & 0.129 & 0.007 & 0.119 & 0.012 \\
\cmidrule(l){2-11}
 & SHerLoc-S \cite{kim2025sherloc}
 & 0.796 & 0.890 & 0.580 & 0.696
 & 0.822 & 0.892 & 0.618 & 0.757 & 0.704 \\
 & SHerLoc \cite{kim2025sherloc}
 & 0.812 & 0.893 & 0.650 & 0.759
 & 0.817 & 0.887 & 0.610 & 0.743 & 0.722 \\
\cmidrule(l){2-11}
\rowcolor{green!20}
 & \textbf{SSD (Ours)}
 & \textbf{0.838} & \textbf{0.907} & \textbf{0.700} & \textbf{0.818}
 & \textbf{0.863} & \textbf{0.948} & \textbf{0.663} & \textbf{0.858} & \textbf{0.766} \\
\midrule

\multirow{6}{*}{\rotatebox{90}{4D}}
 & Autoplace~\cite{cai2022autoplace}
 & 0.799 & 0.967 & 0.725 & 0.945 & 0.812 & 0.986 & 0.619 & 0.901 & 0.738 \\
 & MinkLoc3Dv2~\cite{komorowski2022improving}
 & 0.837 & 0.982 & 0.743 & 0.977
 & 0.725 & 0.981 & 0.619 & 0.963 & 0.735 \\
 & TransLoc4D~\cite{zhang2024transloc4d}
 & 0.833 & 0.970 & 0.804 & 0.976 & 0.801 & 0.991 & 0.676 & 0.940 & 0.779 \\
\cmidrule(l){2-11}
 & SHerLoc-S \cite{kim2025sherloc}
 & 0.866 & 0.945 & 0.803 & 0.925
 & 0.914 & 0.986 & 0.908 & 0.988 & 0.873 \\
 & SHerLoc \cite{kim2025sherloc}
 & 0.904 & 0.961 & 0.843 & 0.949
 & 0.950 & 0.995 & 0.923 & 0.988 & 0.905 \\
\cmidrule(l){2-11}
\rowcolor{green!20}
 & \textbf{SSD (Ours)}
 & \textbf{0.949} & \textbf{0.963} & \textbf{0.942} & \textbf{0.973} & \textbf{0.964} & \textbf{0.996} & \textbf{0.959} & \textbf{0.987} & \textbf{0.954} \\
\bottomrule
\end{tabular}%
}
 \vspace{-1.5em}
\end{table}

\subsection{Qualitative analysis.}

\begin{figure}[t]
    \centering
    \includegraphics[width=\linewidth]{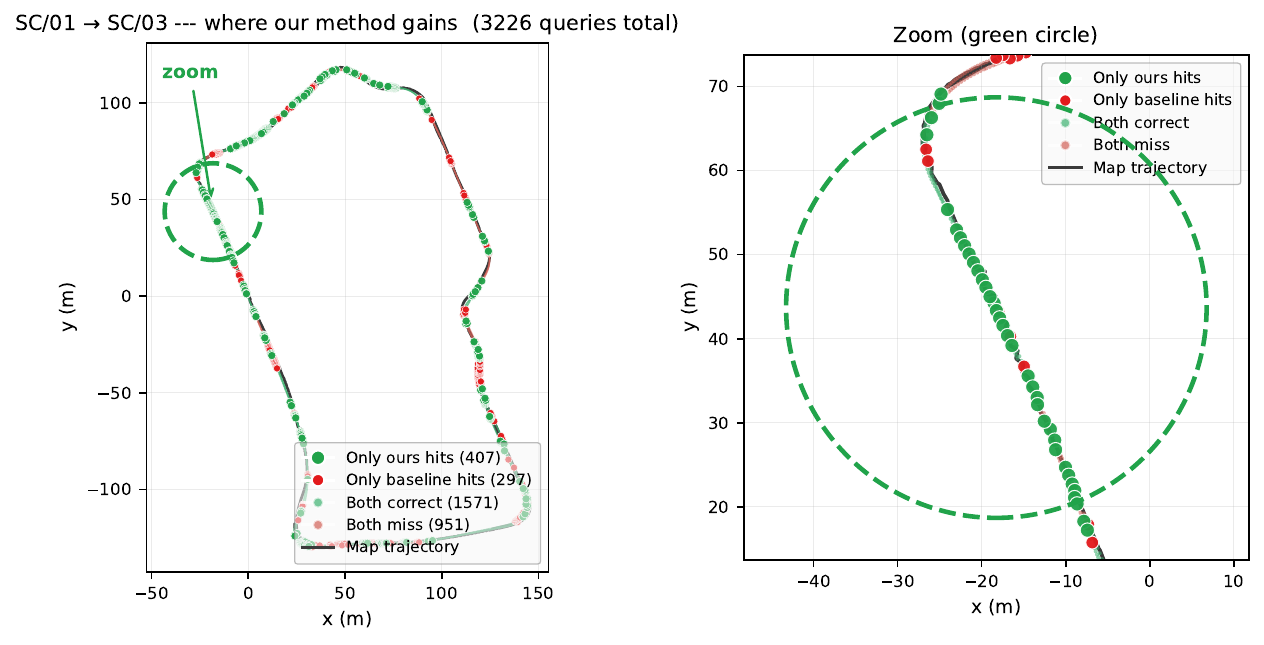}
    \caption{Per-query difference between \textbf{SSD} and \textbf{SHeRLoc} on SC/01$\to$SC/03. \textbf{Green:} SSD-only hits (407); \textbf{Red:} SHeRLoc-only hits (297); \textbf{Dots:} agreement. \textbf{Right:} zoom on the densest win cluster along a curved sub-route.}
    \label{fig:trajectory-diff}
 \vspace{-1.5em}
\end{figure}

Fig.~\ref{fig:trajectory-diff} illustrates SSD's robust retrieval performance on the challenging SC/01$\to$SC/03 multi-session pair, where it recovers 407 queries while incurring only 297 regressions relative to SHeRLoc. As shown, these performance gains are concentrated on a curved sub-route where sparse returns under-constrain place identity. In the most severe failure cases, the baseline predicts locations over $240$\,m away, while SSD keeps it within a $1.2$\,m error margin.

\section{Conclusion}

We presented Spatially-Stratified Distillation (SSD), a cross-modal distillation framework that bridges the sparsity gap between 4D and spinning radar for place recognition. 
Extensive evaluations on the HeRCULES dataset demonstrate that SSD consistently outperforms state-of-the-art baselines across all evaluated single-session and multi-session environments. 
By effectively bridging the structural gap between dense database maps and sparse queries, we hope this work encourages further exploration into asymmetric cross-modal supervision for cost-effective 4D radar localization.

\section{Acknowledgment}
The authors would like to acknowledge the collaboration with Boeing.

\bibliographystyle{ieeetr}
\bibliography{refs.bib}

@article{kim2025sherloc,
  title   = {SHeRLoc: Synchronized Heterogeneous Radar Place Recognition for Cross-Modal Localization},
  author  = {Kim, Hanjun and Jung, Minwoo and Yang, Wooseong and Kim, Ayoung},
  journal = {IEEE Robotics and Automation Letters},
  year    = {2025},
}

@article{kim2025hercules,
  title   = {HeRCULES: A Heterogeneous Radar and LiDAR Dataset for Autonomous Driving},
  author  = {Kim, Hanjun and others},
  journal = {arXiv preprint arXiv:2502.01946},
  year    = {2025},
}

@inproceedings{bang2024radardistill,
  title     = {RadarDistill: Boosting Radar-based Object Detection Performance via Knowledge Distillation from LiDAR Features},
  author    = {Bang, Geonho and Choi, Kwangjin and Kim, Jisong and Kum, Dongsuk and Choi, Jun Won},
  booktitle = {CVPR},
  year      = {2024},
}

@inproceedings{zhang2024transloc4d,
  title     = {TransLoc4D: Transformer-Based 4D Radar Place Recognition},
  author    = {Zhang, Liu and others},
  booktitle = {CVPR},
  year      = {2024},
}

@inproceedings{kim2020mulran,
  title     = {MulRan: Multimodal Range Dataset for Urban Place Recognition},
  author    = {Kim, Giseop and Park, Yeong Sang and Cho, Younghun and Jeong, Jinyong and Kim, Ayoung},
  booktitle = {ICRA},
  pages     = {6246--6253},
  year      = {2020},
}

@inproceedings{jang2023raplace,
  title     = {RaPlace: Place Recognition for Imaging Radar using Radon Transform and Mutable Threshold},
  author    = {Jang, Hyesu and Jung, Minwoo and Kim, Ayoung},
  booktitle = {IROS},
  pages     = {11194--11201},
  year      = {2023},
}

@inproceedings{cai2022autoplace,
  title     = {AutoPlace: Robust Place Recognition with Single-Chip Automotive Radar},
  author    = {Cai, Kaiwen and Wang, Bing and Lu, Chris Xiaoxuan},
  booktitle = {ICRA},
  pages     = {2222--2228},
  year      = {2022},
}

@inproceedings{komorowski2022improving,
  title     = {Improving Point Cloud Based Place Recognition with Ranking-based Loss and Large Batch Training},
  author    = {Komorowski, Jacek},
  booktitle = {ICPR},
  pages     = {3699--3705},
  year      = {2022},
}

@inproceedings{vidanapathirana2022logg3d,
  title={LoGG3D-Net: Locally guided global descriptor learning for 3D place recognition},
  author={Vidanapathirana, Kavisha and Ramezani, Milad and Moghadam, Peyman and Sridharan, Sridha and Fookes, Clinton},
  booktitle={2022 International Conference on Robotics and Automation (ICRA)},
  pages={2215--2221},
  year={2022},
  organization={IEEE}
}

@article{vidanapathirana2023spectral,
  title={Spectral geometric verification: Re-ranking point cloud retrieval for metric localization},
  author={Vidanapathirana, Kavisha and Moghadam, Peyman and Sridharan, Sridha and Fookes, Clinton},
  journal={IEEE Robotics and Automation Letters},
  volume={8},
  number={5},
  pages={2494--2501},
  year={2023},
  publisher={IEEE}
}

@article{hausler2025pair,
  title={Pair-vpr: Place-aware pre-training and contrastive pair classification for visual place recognition with vision transformers},
  author={Hausler, Stephen and Moghadam, Peyman},
  journal={IEEE Robotics and Automation Letters},
  year={2025},
  publisher={IEEE}
}

@article{knights2026wildcross,
  title={WildCross: A Cross-Modal Large Scale Benchmark for Place Recognition and Metric Depth Estimation in Natural Environments},
  author={Knights, Joshua and Reid, Joseph and Roy, Kaushik and Hall, David and Cox, Mark and Moghadam, Peyman},
  journal={arXiv preprint arXiv:2603.01475},
  year={2026}
}

@inproceedings{gadd2024open,
  title={Open-RadVLAD: Fast and robust radar place recognition},
  author={Gadd, Matthew and Newman, Paul},
  booktitle={2024 IEEE Radar Conference (RadarConf24)},
  pages={1--6},
  year={2024},
  organization={IEEE}
}

\end{document}